\documentclass[10pt,twocolumn,letterpaper]{article}

\usepackage{ijcb}
\usepackage{times}
\usepackage{epsfig}
\usepackage{graphicx}
\usepackage{amsmath}
\usepackage{amssymb}
\ijcbfinalcopy

\begin{document}

\title{Fairness on Synthetic Visual and Thermal Mask Images}

\author{Kenneth Lai, Svetlana N. Yanushkevich, Vlad Shmerko\\
	Biometric Technologies Laboratory, Dept. Electrical \& Somputer Engineering, University of Calgary\\
	{\tt\small \{kelai, syanshk, vshmerko\}@ucalgary.ca}
}
\maketitle
\thispagestyle{empty}

\begin{abstract}
In this paper, we study performance and fairness on visual and thermal images and expand the assessment to masked synthetic images.  Using the SpeakingFace and Thermal-Mask dataset, we propose a process to assess fairness on real images and show how the same process can be applied to synthetic images. The resulting process shows a demographic parity difference of 1.59 for random guessing and increases to 5.0 when the recognition performance increases to a precision and recall rate of 99.99\%.  We indicate that inherently biased datasets can deeply impact the fairness of any biometric system.  A primary cause of a biased dataset is the class imbalance due to the data collection process.  To address imbalanced datasets, the classes with fewer samples can be augmented with synthetic images to generate a more balanced dataset resulting in less bias when training a machine learning system. For biometric-enabled systems, fairness is of critical importance, while the related concept of Equity, Diversity, and Inclusion (EDI) is well suited for the generalization of fairness in biometrics, in this paper, we focus on the 3 most common demographic groups age, gender, and ethnicity.
\end{abstract}


\section{Introduction}
Face recognition plays an important task in the ever-growing domains of computer vision and artificial intelligence (AI).  With the adoption of deep machine learning techniques, facial recognition performance has reached extraordinary, a 0.1\% rank one miss rate on a gallery of 12 million individuals \cite{grother2019face}.  An emerging problem in this domain is the vulnerability to biases which results in unfair decisions.  Due to the data-dependent nature of most contemporary machine learning techniques, existing biases in data can bias the underlying algorithms and in some cases may amplify these biases. A biased AI-based decision may lead to unfair treatment such as scenarios in the hiring process \cite{cohen2019efficient}.  These growing concerns encourage the development of a ``fair'' AI system which is critical for the future of AI-based decision-making.

The creation of a ``fair'' system is a multi-stage development process and is dependent on understanding what is bias and how the mitigation of bias can lead to fairness. In \cite{ntoutsi2020bias}, bias is defined as the ``inclination or prejudice of a decision made by an AI system which is for or against one person or group, especially in a way considered to be unfair.''  In \cite{mehrabi2021survey}, fairness is defined as ``the absence of any prejudice or favoritism toward an individual or group based on their inherent or acquired characteristics.'' The two most common biases, gender and racial bias, can manifest in a dataset due to the nature of data collection resulting in over or under-representation of the different demographic groups.  A possible solution is sub-sample over-represented groups while synthetically augmenting under-represented groups.

In this paper, we evaluate the facial recognition performance and fairness metrics for visual and thermal images.  We contrast these results with their synthetic masked counterpart to illustrate how the same fairness analysis can be used to both assess the fairness of real and synthetic images.


\section{Method}
We propose to create a process to evaluate the performance and fairness of a facial recognition system that can be applied to both real and synthetic images as well as visual and thermal modalities.  For the experimental dataset, we choose to use the SpeakingFace \cite{abdrakhmanova2021speakingfaces} and Thermal-Mask \cite{queiroz2021thermal} dataset because of the visual and thermal modalities.  We adopt the demographic parity and equalised odds to assess fairness while using precision, recall, and f1-score for performance evaluation.  The proposed facial recognition process uses a simple 2-Block and 3-Block convolutional neural network.

\subsection{Dataset}
\begin{figure}[!hbt]
	\hspace{-2.5mm}
	\begin{tabular}{cccc}
		\includegraphics[width=.1\textwidth]{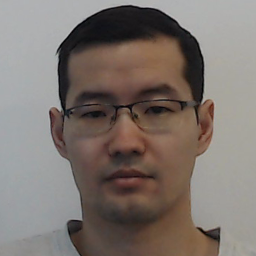} &
		\includegraphics[width=.1\textwidth]{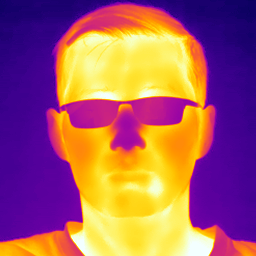} &
		\includegraphics[width=.1\textwidth]{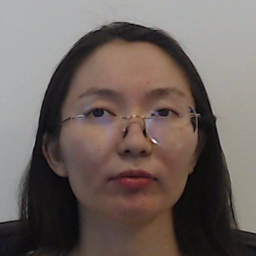} &
		\includegraphics[width=.1\textwidth]{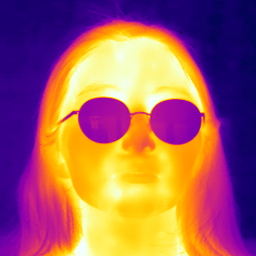} \\
		(a)	&	(b)	&	(c)	&	(d)	\\
	\end{tabular}
	\caption{Example images from the SpeakingFace Dataset: (a) subject 1 visual, (b)  subject 1 thermal, (c) subject 16 visual, (d)  subject 16 thermal.}
	\label{fig:speaking}
\end{figure}

SpeakingFaces Dataset \cite{abdrakhmanova2021speakingfaces}: a large-scale multimodal dataset that combines thermal, visual, and audio data streams. It includes data from 142 subjects, with a gender balance of 68 female and 74 male participants, with ages ranging from 20 to 65 years with an average of 31 years. With approximately 4.6 million images collected in both the visible and thermal spectra, each of the 142 subjects has nine different head positions and each position with 900 frames acquired in 2 trials.  Fig. \ref{fig:speaking} shows the thermal and visual images of 2 different subjects.

\begin{figure}[!hbt]
	\hspace{-2.5mm}
	\begin{tabular}{cccc}
		\includegraphics[width=.1\textwidth]{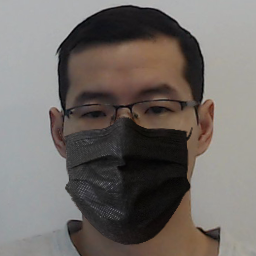} &
		\includegraphics[width=.1\textwidth]{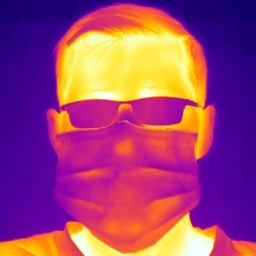} &
		\includegraphics[width=.1\textwidth]{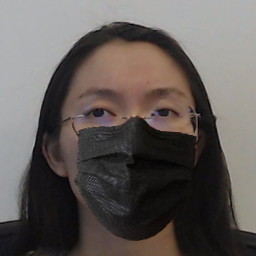} &
		\includegraphics[width=.1\textwidth]{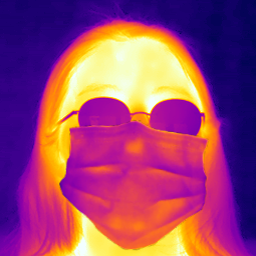} \\
		(a)	&	(b)	&	(c)	&	(d)	\\
	\end{tabular}
	\caption{Example images from the Thermal-Mask Dataset: (a) subject 1 visual, (b)  subject 1 thermal, (c) subject 16 visual, (d) subject 16 thermal.}
	\label{fig:thermal}
\end{figure}

Thermal-Mask dataset \cite{queiroz2021thermal}: A synthetic mask dataset created using the SpeakingFaces Dataset.  This dataset consists of 80 subjects with a total of 84,920 synthetic masked visual and thermal images.  The images in this dataset are cropped and aligned with their SpeakingFaces counterpart and have a pixel resolution of 256x256.  Fig. \ref{fig:thermal} shows the thermal and visual mask images of 2 different subjects.

\subsection{Performance Metrics}
In this paper, the performance of the machine learning models is measured in terms of Precision, Recall, and F1-Score (F1) defined in Eq. \ref{eq:prec}, \ref{eq:rec}, \ref{eq:f1}.

\begin{equation}\label{eq:prec}
	\text{Precision} = \frac{TP}{TP+FP}
\end{equation}
\begin{equation}\label{eq:rec}
	\text{Recall} = \frac{TP}{TP+FN}
\end{equation}
\begin{equation}\label{eq:f1}
	\text{F1} = \frac{2*\text{Precision}*\text{Recall}}{\text{Precision}+\text{Recall}}=\frac{2TP}{2TP+FP+FN}
\end{equation}

where $TP$ (True Positives) represents correct recognition of the genuine user, $TN$ (True Negatives) represents the correct recognition of imposters, $FP$ (False Positives) represents incorrect recognition of imposters as the genuine user, and $FN$ (False Negatives) represents the incorrect recognition of the genuine user as imposters.

\subsection{Fairness Metrics}
In this paper, we evaluate the fairness of the facial recognition system using the demographic parity difference \cite{[Dwork-2012]} and equalized odds difference \cite{hardt2016equality}.

Demographic parity, also known as statistical parity, states that for each protected group, their positive rate should be similar. A system satisfies statistical parity if its prediction is statistically independent of the demographic group.  This system can be represented as:
\begin{equation}
	Pr(\hat{y}|D=a) =  Pr(\hat{y}|D=b) =  Pr(\hat{y}|D=z)
\end{equation}
where $\hat{y}$ represents the predictor, $D$ is the demographic group (e.g. gender, ethnicity, etc), and $a,b,...,z\in D$ are the classes (e.g. male and female) in demographic group $D$.

Demographic parity difference (DPD) is defined as the difference in positive rate between the largest and smallest demographic group. A DPD of 0 represents that all demographic groups have the same positive rate.
\begin{equation}\label{eq:dpd}
	DPD = Pr(\hat{y}|D=l)-Pr(\hat{y}|D=s)
\end{equation}
where $l, s \in D$ represents the largest and smallest class in the demographic group $D$, respectively. 
 
Equalized odds state that the true positive rate and false positive rate across each protected group should be similar.  A system satisfies equal opportunity if its prediction is conditionally independent of the protected group. This system can be represented as:
 \begin{equation}
 	Pr(\hat{y}|y,D=a) =  Pr(\hat{y}|y,D=z)
 \end{equation}
where $\hat{y}$ represents the predictor, $y$ represents conditionally positive outcome, and $a,b,...,z\in D$ are the classes in protected group $D$.

The equalized odds difference (EOD) is defined as the larger of the two: true positive rate difference (TPD) and false positive rate difference (FPD).  The TPD is defined as the difference in true positive rate between the largest and smallest demographic group.  Similarly, the FPD is the difference in false positive rate between the largest and smallest demographic group.  An EOD of 0 represents that all demographic groups have the same true positive, true negative, false positive, and false negative rates.
\begin{equation}
	TPD = Pr(\hat{y}|y,D=l) - Pr(\hat{y}|y,D=s)
\end{equation}
\begin{equation}
	FPD = Pr(\hat{y}|y',D=l) - Pr(\hat{y}|y',D=s)
\end{equation}
\begin{equation}\label{eq:eod}
	EOD = Max(TPD,FPD)
\end{equation}
where $y'$ represents a conditionally negative outcome and $l, s \in D$ represents the largest and smallest class in the demographic group $D$, respectively.

\subsection{Convolutional Neural Network}
In this paper, we choose to use two simplistic deep convolutional neural networks to evaluate the overall facial recognition performance and fairness metric.  Table \ref{tab:cnn} shows the CNN architectures used in this paper.  Both CNNs were using the Adam optimizer with default parameters: learning rate $\alpha=0.001$, $\beta_1=0.9$, and $\beta_2=0.999$.  For each subject, 10\% of the images were used for training and the remaining 90\% were used for testing.  The networks were trained with a batch size of 32 for a total of 10 epochs.  This extreme partition of training/testing sets accompanied by low epoch count was chosen to evaluate the role of imbalanced data on facial recognition, specifically assessing fairness in the dataset.  The purpose of the experiment is not to maximize facial recognition performance but to observe the change in fairness between real and synthetic images.

\begin{table}[!htb]
	\centering
	\caption{2-Block and 3-Block CNN Architecture}\label{tab:cnn}
	\begin{tabular}{c|c}
			2-block CNN	&	3-block CNN	\\
			\hline
		Input 256x256x3 &	Input 256x256x3\\
		\hline
		64 Conv2D 3x3	&	64 Conv2D 3x3	\\
		64 Conv2D 3x3	&	64 Conv2D 3x3	\\
		Max Pooling	&	Max Pooling	\\
		Batch Normalization	&	Batch Normalization	\\
		\hline
		128 Conv2D 3x3	&	128 Conv2D 3x3	\\
		128 Conv2D 3x3	&	128 Conv2D 3x3	\\
		Max Pooling	&	Max Pooling	\\
		Batch Normalization	&	Batch Normalization	\\
		\hline	
				&	256 Conv2D 3x3\\
				&	256 Conv2D 3x3\\
				&	Max Pooling\\
				&	Batch Normalization\\
		\hline	
		\multicolumn{2}{c}{Global Average Pooling}	\\		
		\multicolumn{2}{c}{Fully-Connected}	\\		
		\multicolumn{2}{c}{Softmax Classification}	\\
	\end{tabular}
\end{table}

\section{Experimental Results}
The experimental study involves the use of the two simple 2-block and 3-block convolutional neural networks to perform facial recognition.  The performance (precision, recall, and F1-Score) and fairness are assessed using the test set which consists of 90\% (38222 images) of the images.

\subsection{Performance Across Groups}
In this paper, we evaluate the performance and fairness on 3 different demographic groups: gender, ethnicity, and age.  For gender, we separate the dataset into binary male/female classes based on the gender labels provided in the dataset. For ethnicity, we divide into 3 categories: A, B, and C, based on the ethnicity labels provided for each subject.  Lastly, for age, we split the dataset into 4 groups: $<$25, 25 to 30, 31 to 35, and $>$ 35, based on the reported age of each subject. The probability distributions for each demographic group is as follows:
	\begin{footnotesize}
	\begin{center}
		\begin{tabular}{cc}
			\begin{tabular}{cc}
				\multicolumn{2}{c}{Gender} \\
				 Male & Female \\
				\hline
				0.5625 & 0.4375 \\
			\end{tabular} & 
			\begin{tabular}{ccc}
				\multicolumn{3}{c}{Ethnicity} \\
				 A & B & C \\
				\hline
				 0.750 & 0.0375	&.2125	\\
			\end{tabular} \\
		
		\multicolumn{2}{c}{\begin{tabular}{cccc}
			\multicolumn{4}{c}{Age} \\
			$<$25 & 25-30 & 31-35 & $>$ 35 \\
			\hline
			0.3875 & 0.3250	&0.1250	&0.1625	\\
		\end{tabular} }
		\end{tabular} 
	\end{center}	
\end{footnotesize}

We can see from the prior probability distribution for each demographic group that the dataset is not balanced, that is the ratio of male-to-female or A-to-B-to-C is not equal distribution.  When an imbalanced dataset is used for training a CNN, it can lead to a biased network.  An example of the performance of a biased network is shown in Table \ref{tab:bias}.  The 2-block CNN is used as the recognition model with SpeakingFace (un-mask) and thermal-mask (mask) datasets used for evaluation. The rows represent the performance based on the different demographic groups such as gender, age, and ethnicity. For this experiment, we show the performance in terms of precision, recall, and F1-score measured for visual or thermal images as well as real (un-mask) or synthetic (masked) images.  For example, the second row in Table \ref{tab:bias} shows the performance of subjects with an age of 32.  This group shows high performance for thermal images regardless of the real or synthetic nature.  An interesting disparity is shown when comparing the real and synthetic performance of the visual images using recall and f1-scores.  The masked recall rate for the visual image is 30.45\% while the un-masked recall rate is 99.76\%.

\begin{table*}[!htb]
	\centering
	\begin{footnotesize}
		\caption{2-Block CNN Facial Recognition Performance in terms of Precision, Recall, and F1-Score.}\label{tab:bias}
		\begin{tabular}{l|rr|rr|rr||rr|rr|rr}	
			& 	\multicolumn{6}{c||}{Visual} & \multicolumn{6}{c}{Thermal} \\
			\cline{2-13}
			&	\multicolumn{2}{c|}{Precision}	&	\multicolumn{2}{c|}{Recall}	&	\multicolumn{2}{c||}{F1-Score} &	\multicolumn{2}{c|}{Precision}	&	\multicolumn{2}{c|}{Recall}	&	\multicolumn{2}{c}{F1-Score}\\
			&	Mask & Un-Mask	&	Mask & Un-Mask	&	Mask & Un-Mask &	Mask & Un-Mask	&	Mask & Un-Mask	&	Mask & Un-Mask\\
			\hline
			\hline
			baseline	&	70.94	&	88.10	&	53.93	&	83.01	&	50.68	&	82.85	&		69.96	&	85.82	&	68.92	&	85.66	&	68.92	&	85.66	\\
			\hline
			age:32	&	100.00	&	97.98	&	30.45	&	99.79	&	46.69	&	98.88	&		100.00	&	100.00	&	91.96	&	97.01	&	91.96	&	97.01	\\
			age:37	&	77.67	&	100.00	&	59.52	&	80.96	&	58.61	&	89.32	&		77.48	&	97.34	&	84.06	&	93.73	&	84.06	&	93.73	\\
			age:29	&	87.87	&	95.33	&	58.19	&	88.73	&	64.08	&	91.11	&		76.74	&	88.16	&	74.04	&	89.65	&	74.04	&	89.65	\\
			age:27	&	57.05	&	70.56	&	55.44	&	79.66	&	28.09	&	68.75	&		76.57	&	89.62	&	65.72	&	86.71	&	65.72	&	86.71	\\
			age:24	&	84.53	&	83.31	&	76.61	&	89.03	&	76.68	&	85.41	&		82.30	&	93.28	&	79.37	&	94.19	&	79.37	&	94.19	\\
			age:25	&	72.97	&	84.13	&	44.20	&	85.60	&	46.53	&	80.91	&		72.09	&	93.76	&	68.41	&	92.58	&	68.41	&	92.58	\\
			age:21	&	78.41	&	92.48	&	65.50	&	73.72	&	60.75	&	79.84	&		50.76	&	76.53	&	56.18	&	80.58	&	56.18	&	80.58	\\
			age:22	&	46.41	&	88.56	&	38.81	&	78.13	&	38.99	&	82.76	&		74.76	&	87.02	&	79.66	&	84.49	&	79.66	&	84.49	\\
			age:23	&	65.48	&	82.46	&	49.04	&	86.64	&	38.67	&	81.52	&		57.60	&	77.60	&	64.53	&	83.47	&	64.53	&	83.47	\\
			age:33	&	93.17	&	98.55	&	65.95	&	75.41	&	73.91	&	85.28	&		67.39	&	99.79	&	55.06	&	92.02	&	55.06	&	92.02	\\
			age:35	&	54.96	&	68.65	&	91.15	&	99.59	&	68.58	&	81.28	&		87.04	&	96.91	&	93.07	&	96.62	&	93.07	&	96.62	\\
			age:30	&	80.27	&	87.98	&	56.84	&	92.90	&	54.56	&	90.02	&		67.53	&	70.72	&	74.99	&	79.71	&	74.99	&	79.71	\\
			age:57	&	77.94	&	100.00	&	66.87	&	98.56	&	71.98	&	99.27	&		100.00	&	100.00	&	67.59	&	98.28	&	67.59	&	98.28	\\
			age:36	&	61.58	&	87.67	&	77.20	&	97.05	&	65.17	&	91.95	&		66.35	&	65.99	&	68.41	&	65.79	&	68.41	&	65.79	\\
			age:28	&	66.14	&	83.04	&	48.08	&	81.28	&	53.66	&	80.84	&		51.30	&	66.74	&	64.89	&	79.25	&	64.89	&	79.25	\\
			age:26	&	64.22	&	99.52	&	52.61	&	77.85	&	57.78	&	83.17	&		91.50	&	84.64	&	92.48	&	91.00	&	92.48	&	91.00	\\
			age:41	&	34.28	&	89.76	&	26.03	&	86.52	&	29.59	&	88.11	&		67.59	&	91.67	&	60.16	&	93.43	&	60.16	&	93.43	\\
			age:20	&	67.43	&	87.48	&	46.34	&	72.14	&	46.40	&	73.64	&		72.92	&	91.23	&	59.64	&	78.30	&	59.64	&	78.30	\\
			age:45	&	96.70	&	97.92	&	20.09	&	75.11	&	33.27	&	85.01	&		85.84	&	97.95	&	39.19	&	95.76	&	39.19	&	95.76	\\
			age:34	&	70.07	&	91.43	&	32.92	&	75.62	&	38.36	&	81.12	&		69.10	&	91.79	&	64.66	&	75.01	&	64.66	&	75.01	\\
			age:31	&	95.67	&	100.00	&	37.41	&	87.47	&	51.99	&	93.21	&		67.85	&	97.06	&	78.75	&	96.54	&	78.75	&	96.54	\\
			age:40	&	90.23	&	93.48	&	47.53	&	94.44	&	62.26	&	93.96	&		94.44	&	98.56	&	86.36	&	99.27	&	86.36	&	99.27	\\
			age:39	&	91.47	&	93.00	&	94.91	&	100.00	&	93.16	&	96.38	&		96.02	&	96.90	&	97.86	&	96.48	&	97.86	&	96.48	\\
			age:46	&	35.85	&	100.00	&	95.82	&	81.37	&	52.17	&	89.73	&		78.33	&	75.67	&	80.31	&	74.81	&	80.31	&	74.81	\\
			\hline
			gender:Male	&	70.31	&	87.38	&	55.70	&	82.97	&	50.95	&	83.27	&		68.77	&	88.01	&	66.26	&	88.22	&	66.26	&	88.22	\\
			gender:Female	&	71.76	&	89.03	&	51.66	&	83.07	&	50.34	&	82.32	&		71.50	&	83.02	&	72.35	&	82.36	&	72.35	&	82.36	\\
			\hline
			ethnicity:A	&	70.97	&	86.10	&	50.66	&	81.96	&	47.32	&	81.16	&		66.53	&	84.34	&	66.09	&	84.48	&	66.09	&	84.48	\\
			ethnicity:B	&	75.95	&	86.97	&	66.87	&	84.77	&	57.17	&	83.93	&		89.71	&	97.05	&	92.36	&	97.85	&	92.36	&	97.85	\\
			ethnicity:C	&	69.96	&	95.33	&	63.18	&	86.42	&	61.41	&	88.64	&		78.59	&	89.08	&	74.80	&	87.65	&	74.80	&	87.65	\\
		\end{tabular}
	\end{footnotesize}
\end{table*}

\subsection{Fairness Across Real and Synthetic Images}
Table \ref{tab:performance_fair} shows the performance and fairness metrics on the two CNNs.  Given the same hyperparameters used for training both CNNs, the 3-Block CNN greatly outperforms the 2-Block CNN.  The 3-Block CNN achieves near-perfect recognition performance on the thermal images with a slight decrease for the visual images.  The demographic parity difference (DPD) and equalized odds difference (EOD) is calculated based on Eq. \ref{eq:dpd} and \ref{eq:eod}, respectively.  The rows in the table represent the data used for experiments and the column represents the different performance/fairness metrics used.  We can see that as the recognition performance approaches 100\% the DPD approaches 5, 2.5, and 3.75 for age, gender, and ethnicity, respectively.  The approached value is the quotient of the number of classes in the demographic group and the number of subjects to be recognized.  For example, using gender, the approached value is calculated as $\frac{2 classes}{80 subjects}\times100 = \fbox{2.50}$.  As we decrease the performance of recognition, either by reducing the model learning capacity or increasing noise in an image, we can see a decrease in DPD and EOD.  The last row of the table simulates the performance of random guessing which is equivalent to $1/80=\fbox{1.25}$ if the number of samples per subject is the same; however, since there are an unbalanced number of images per subject, the random performance is approximately \fbox{1.32}.

\begin{table*}[!htb]
	\centering
	\begin{footnotesize}
		\caption{Facial Recognition Performance and Fairness Evaluation}\label{tab:performance_fair}
		\begin{tabular}{l|rrr|rrr|rrr}
		&	\multicolumn{3}{c|}{Performance}		&	\multicolumn{3}{c|}{DPD}	&	\multicolumn{3}{c}{EOD}	\\
				& \multicolumn{1}{c}{Precision}	&	\multicolumn{1}{c}{Recall}	&	\multicolumn{1}{c|}{F1-Score}	&		\multicolumn{1}{c}{Age}	& 	\multicolumn{1}{c}{Gender}	&		\multicolumn{1}{c|}{Ethnicity}	&		\multicolumn{1}{c}{Age}	& 	\multicolumn{1}{c}{Gender}	&		\multicolumn{1}{c}{Ethnicity}	\\
\hline
\hline																					
\multicolumn{10}{c}{2-Block CNN}\\
Mask-Visual	&	70.94	&	53.93	&	50.68	&	3.33	&	2.04	&	3.26	&	53.93	&	53.93	&	53.93	\\
Normal-Visual	&	88.10	&	83.01	&	82.85	&	4.45	&	2.33	&	3.61	&	83.01	&	83.01	&	83.01	\\
Mask-Thermal	&	82.20	&	69.96	&	68.92	&	3.97	&	2.21	&	3.38	&	69.96	&	69.96	&	69.96	\\
Normal-Thermal	&	89.57	&	85.82	&	85.66	&	4.56	&	2.43	&	3.58	&	85.82	&	85.82	&	85.82	\\
%
%
\hline
\multicolumn{10}{c}{3-Block CNN}\\
Mask-Visual	&	99.58	&	99.54	&	99.54	&	4.98	&	2.50	&	3.75	&	99.54	&	99.54	&	99.54	\\
Normal-Visual	&	99.89	&	99.87	&	99.88	&	4.99	&	2.50	&	3.75	&	99.87	&	99.87	&	99.87	\\
Mask-Thermal	&	99.99	&	99.99	&	99.99	&	5.00	&	2.50	&	3.75	&	99.99	&	99.99	&	99.99	\\
Normal-Thermal	&	99.99	&	99.99	&	99.99	&	5.00	&	2.50	&	3.75	&	99.99	&	99.99	&	99.99	\\

\hline
Random-Guess	&	1.32	&	1.33	&	1.32	&	1.59	&	0.10	&	0.29	&	1.76	&	1.33	&	1.34	\\
		\end{tabular}
	\end{footnotesize}
\end{table*}

\begin{figure*}[!ht]
	\begin{center}
		\begin{tabular}{cc}	
		\includegraphics[width=0.4\textwidth]{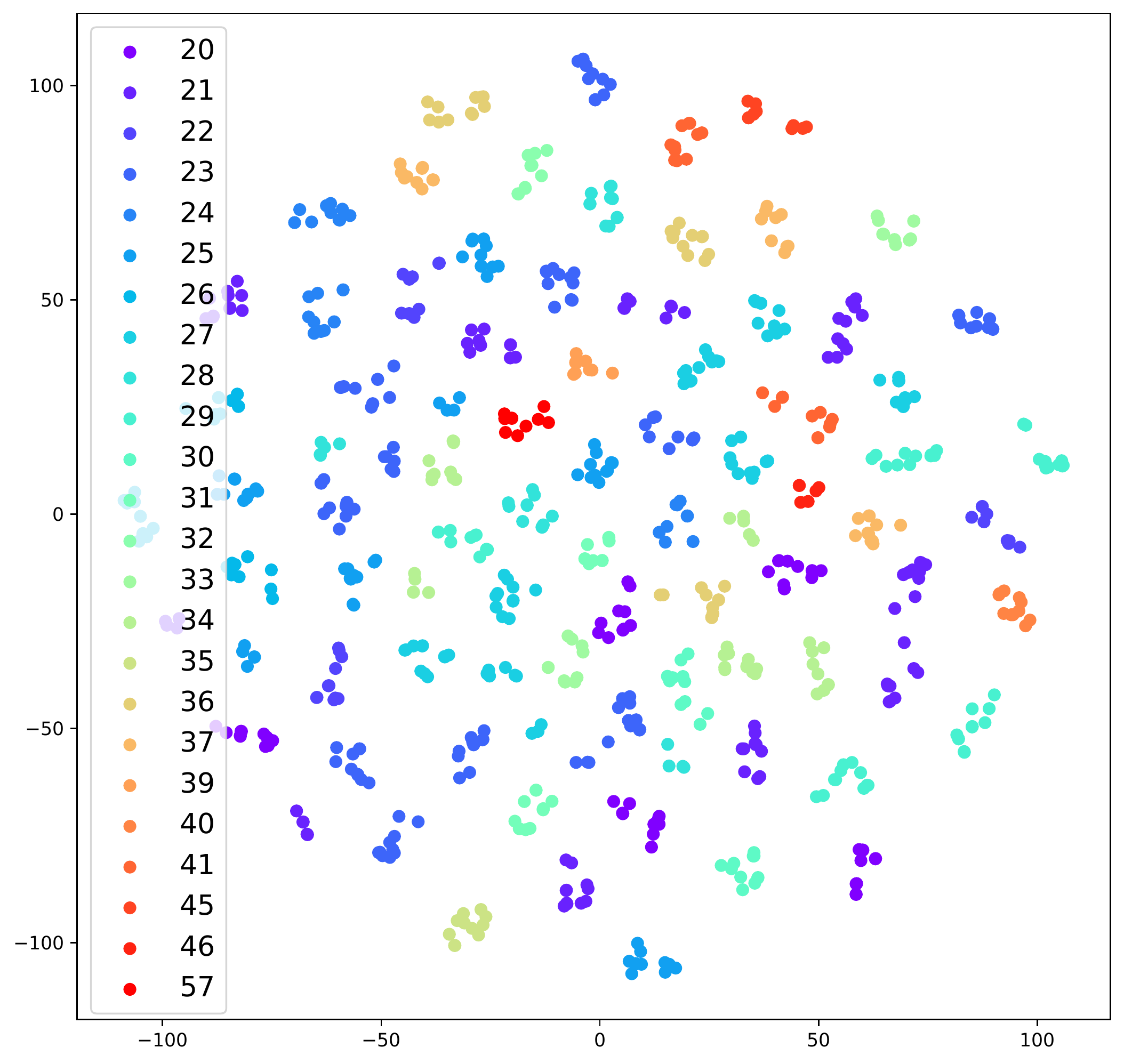} &
		\includegraphics[width=0.4\textwidth]{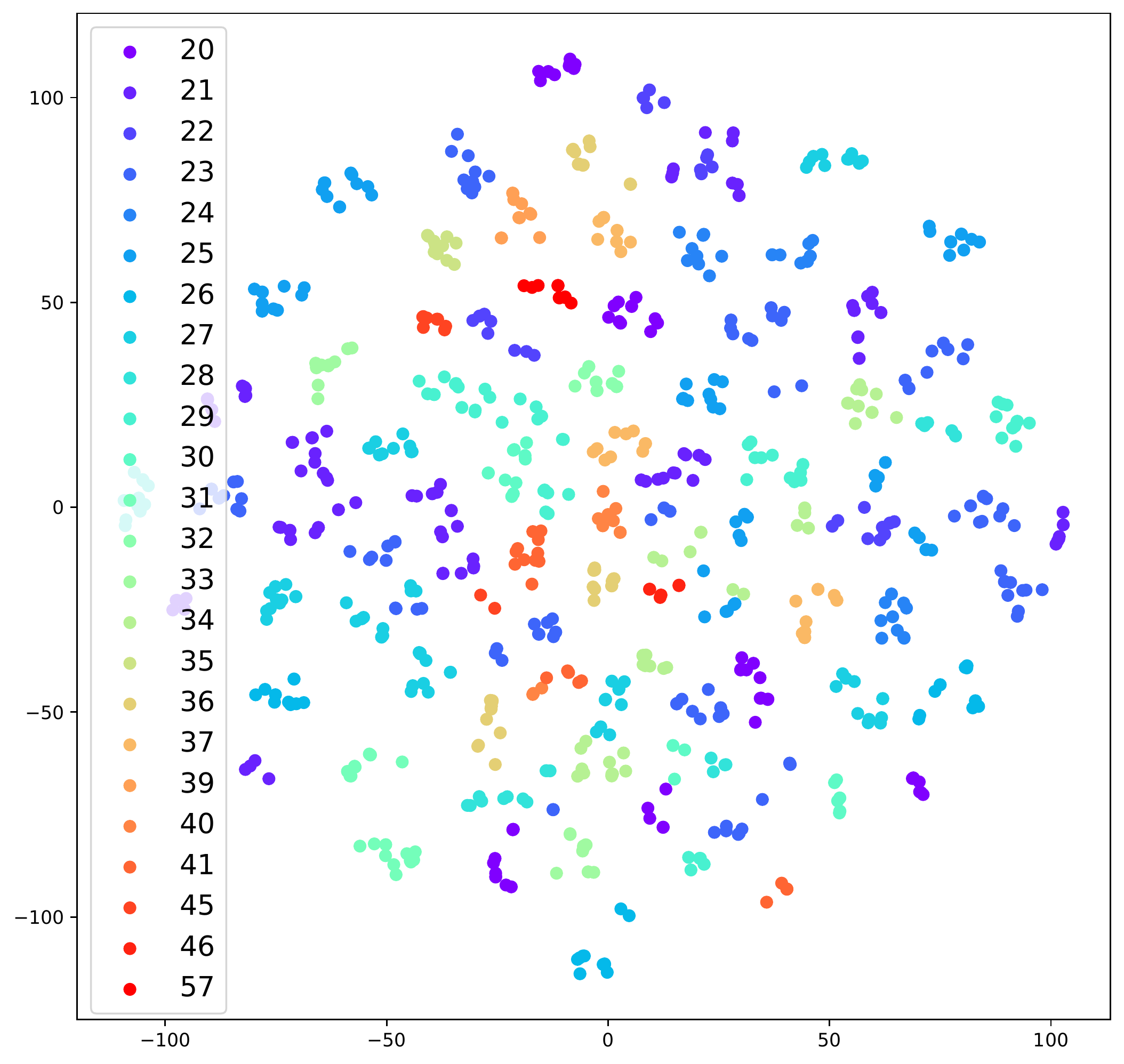} \\
		(a) & (b) \\
		\includegraphics[width=0.4\textwidth]{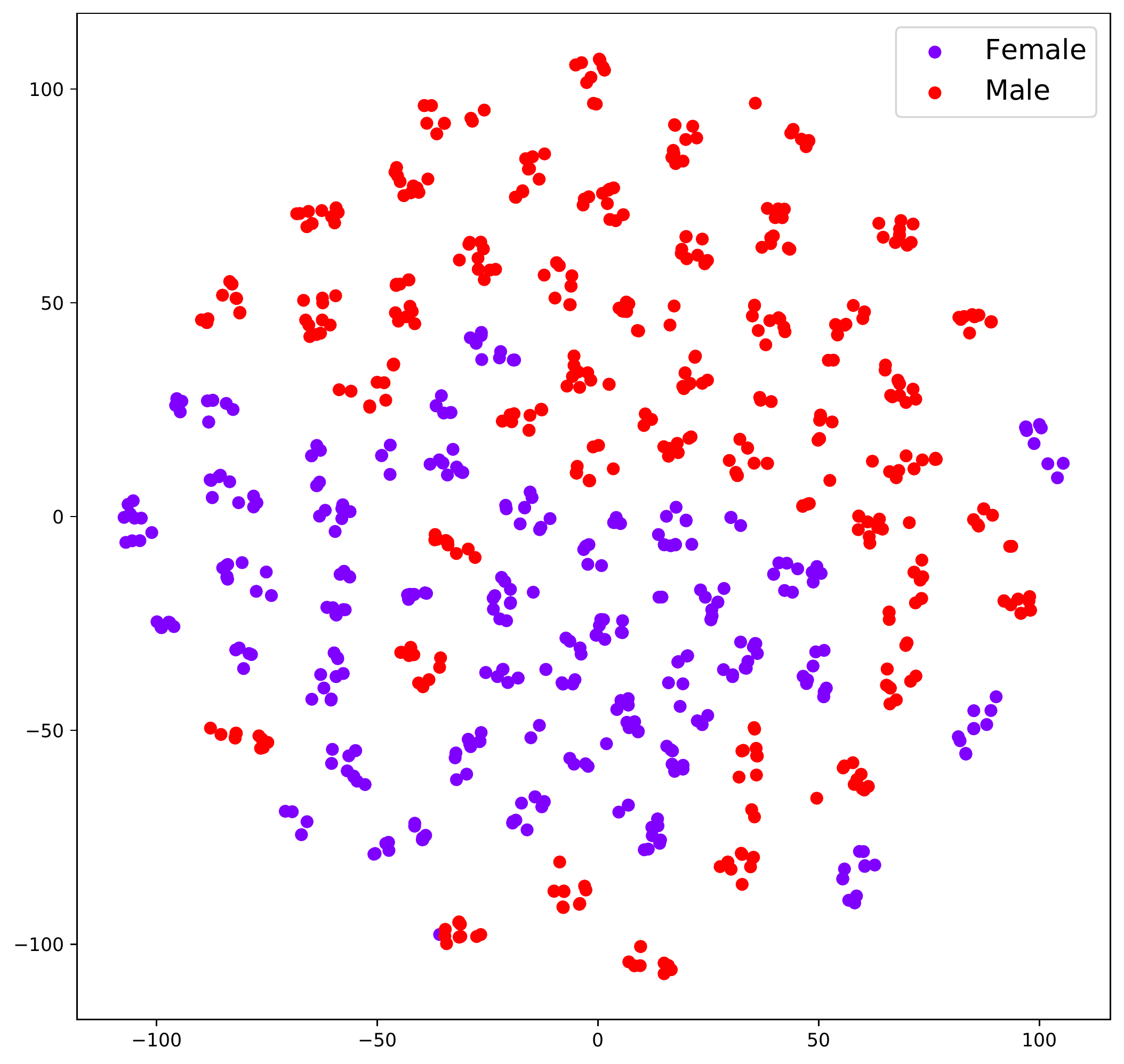} &
		\includegraphics[width=0.4\textwidth]{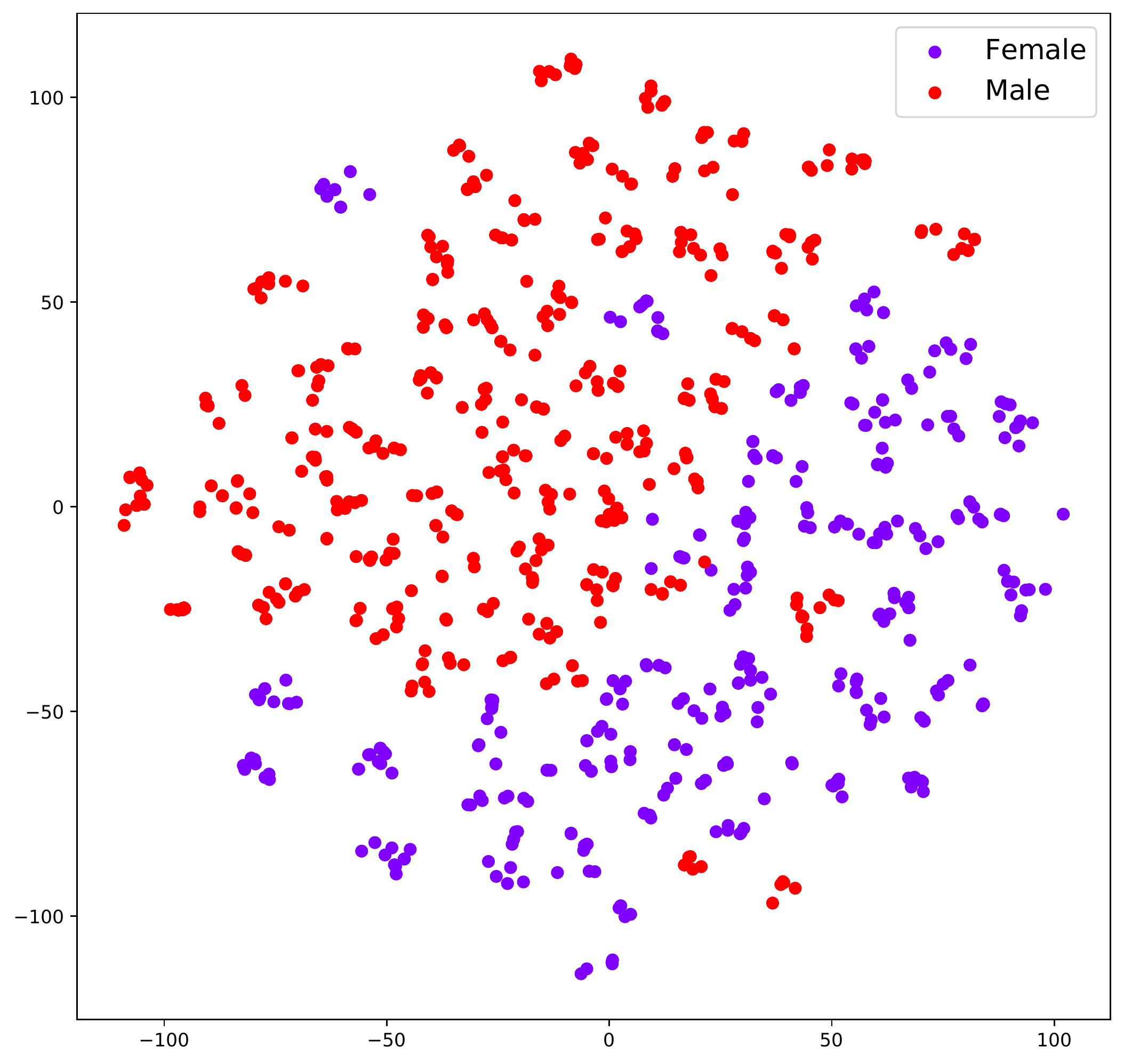} \\
		(c) & (d) \\
		\includegraphics[width=0.4\textwidth]{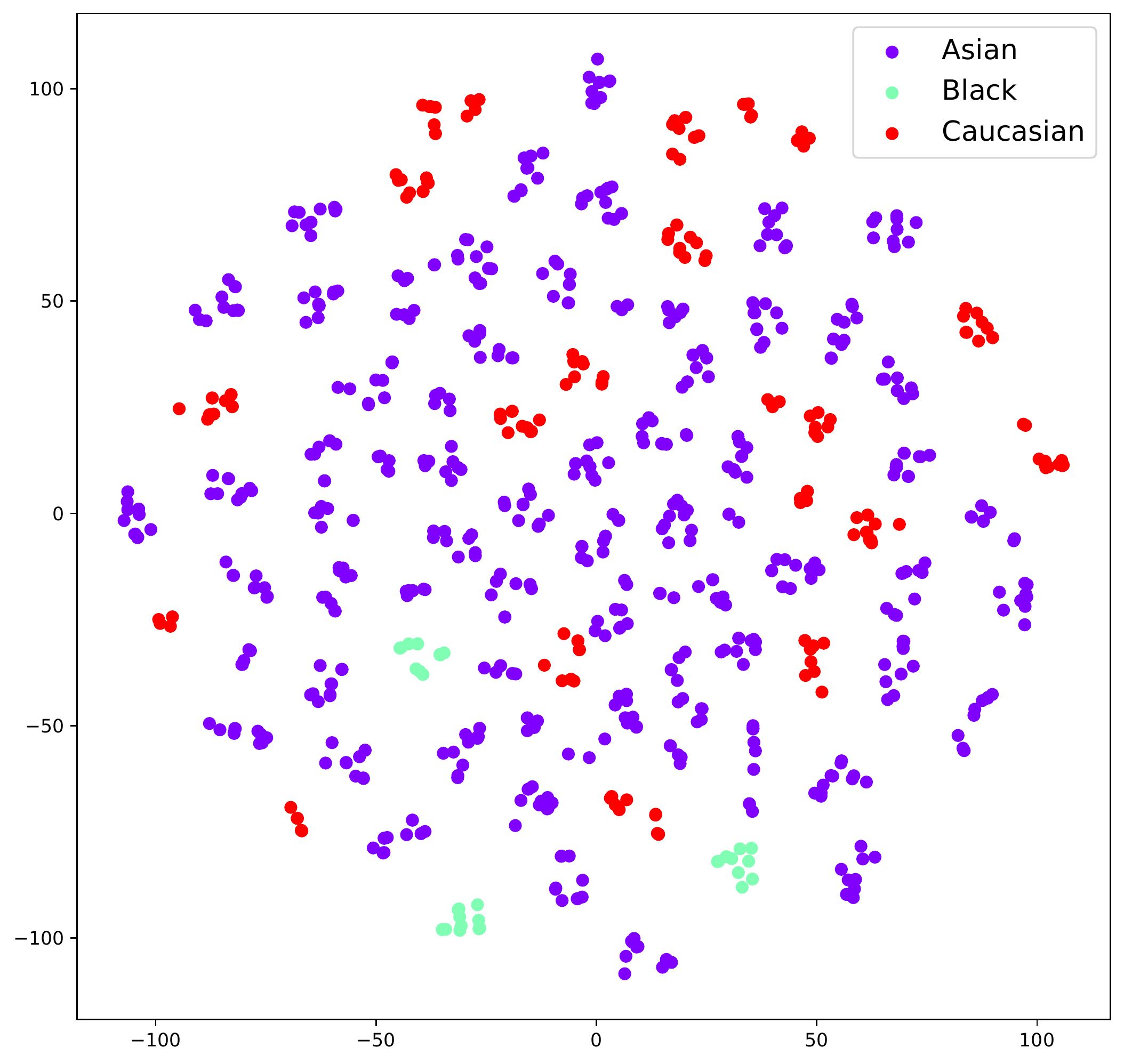} &
		\includegraphics[width=0.4\textwidth]{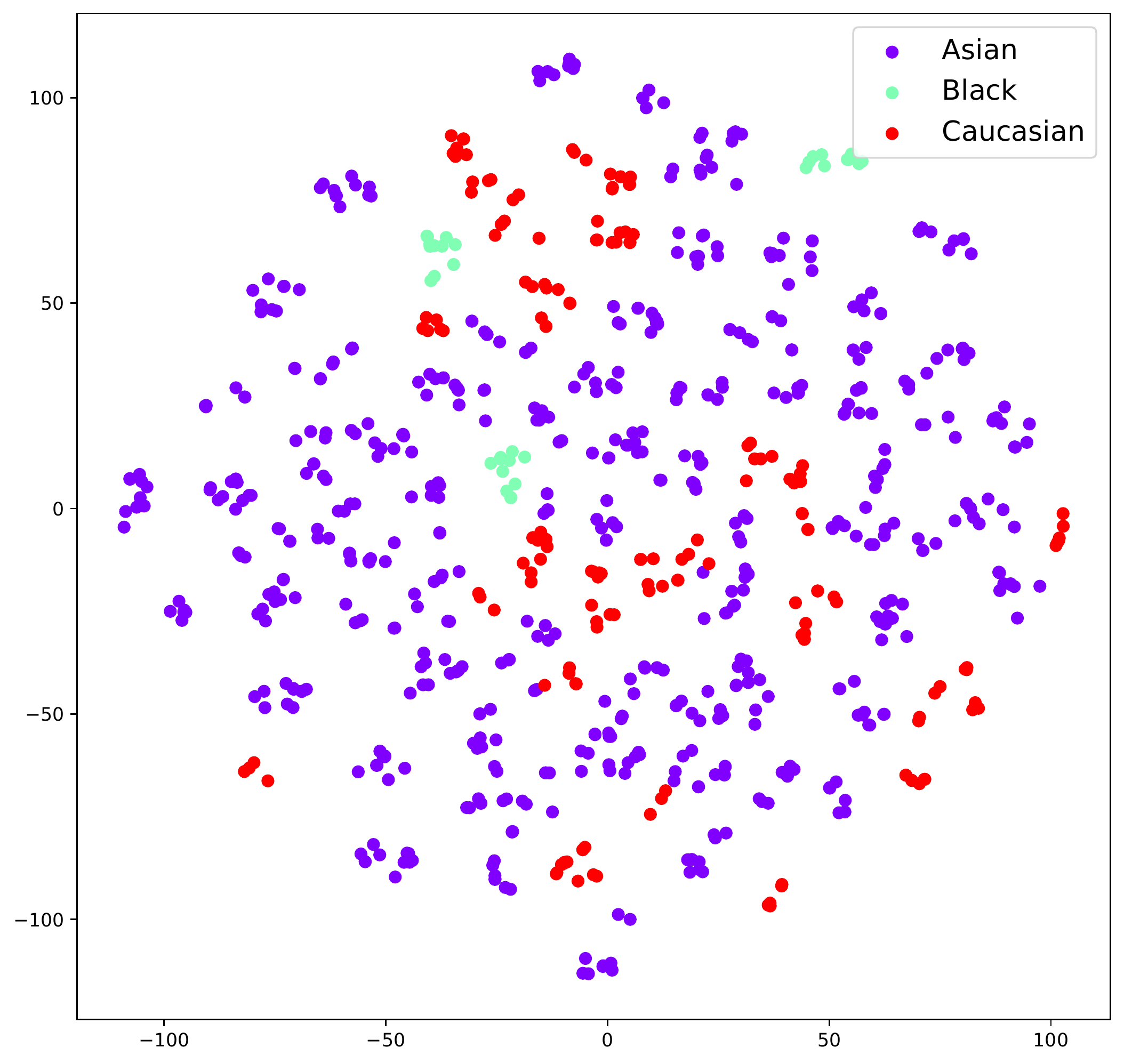} \\
		(e) & (f) \\
		\end{tabular}
	\end{center}
	\caption{The t-SNE visualization of the features extracted using the CNNs: (a) age-based thermal mask plot, (b) age-based thermal un-mask plot,  (c) gender-based thermal mask plot, (d) gender-based thermal un-mask plot, (e) ethnicity-based thermal mask plot, (f) ethnicity-based thermal un-mask plot.}
	\label{fig:tsne-thermal}
\end{figure*}

Fig. \ref{fig:tsne-thermal} shows the t-SNE visualization of the features extracted using the CNNs encoded into a 2D map.  Each point in the t-SNE plots represents an embedded  image.  Multiple points form a cluster that represents a demographic group.  The combination of different demographic groups forms the characteristics of a subject which can be used for identification. For example, taking the bottom  left cluster ($x~-30, y~-90$) indicates an age of 35, Male gender, and Black ethnicity. When combined, these characteristics indicate the person is subject 23.  This process can be applied to both real and synthetic images to identify individuals using the cluster and demographic data.

\section{Conclusions}
Our study addresses the problem of bias and how it can play a role in impacting fairness in a biometric system, specifically a facial recognition system.  Bias can come from a variety of sources, in this paper, we explore how an imbalanced dataset can contain dangerously biased cohorts in the form of demographic groups such as gender, ethnicity, and age.  These biases can deeply influence the machine learning algorithm to make unfair decisions.  In this paper, we show how the same process of evaluation fairness on real images can be replicated on synthetic images.  The evaluation shows that fairness is more correlated to the performance of the system than whether or not the images are synthetic.  As the performance increases, the demographic parity difference also increases proportionally to the number of classes in the demographic group.  Given a simple 3-Block CNN with a precision and recall rate of 99.99\%, the DPD for age, gender, and ethnicity is reported as 5, 2.5, and 3.75, respectively.

A future application is to build a combined real and synthetic dataset where synthetic images are used to augment classes with few samples to create an overall more balanced dataset.

\section*{Acknowledgment}
This research was partially supported by the Natural Sciences and Engineering Research Council Canada (NSERC SPG grant ``Biometric-Enabled Identity Management for Safe and Secure Cities'').  This work was partially supported by Natural Sciences and Engineering Research Council of Canada through Discovery Grant ``Biometric Intelligent Interfaces''.

{\small
\bibliographystyle{IEEEtran}
\bibliography{bib}
}

\end{document}